# Detecting Structural Irregularity in Electronic Dictionaries Using Language Modeling

**Paul Rodrigues, David Zajic, David Doermann, Michael Bloodgood, Peng Ye**

University of Maryland
College Park, MD
E-mail: {prr, dmzajic, doermann, meb, pengyu}@umd.edu

## Abstract

Dictionaries are often developed using tools that save to Extensible Markup Language (XML)-based standards. These standards often allow high-level repeating elements to represent lexical entries, and utilize descendants of these repeating elements to represent the structure within each lexical entry, in the form of an XML tree. In many cases, dictionaries are published that have errors and inconsistencies that are expensive to find manually. This paper discusses a method for dictionary writers to quickly audit structural regularity across entries in a dictionary by using statistical language modeling. The approach learns the patterns of XML nodes that could occur within an XML tree, and then calculates the probability of each XML tree in the dictionary against these patterns to look for entries that diverge from the norm.

**Keywords**: anomaly detection, error correction, dictionaries

## 1. Introduction

Many dictionaries today are developed using tools that save to Extensible Markup Language (XML)-based standards, such as the Lexical Markup Framework (LMF) (Francopoulo et al., 2007), the Lexicon Interchange FormaT (LIFT) (Hosken, 2009), or the Text Encoding Initiative (TEI) (Burnard & Bauman, 2007). Often, these standards allow high-level repeating elements to represent lexical entries and utilize descendants of these repeating elements to represent the structure of each lexical entry, in the form of an XML tree.

This paper presents a method to audit the structural regularity across all the entries in a dictionary, automatically. This approach uses statistical language modeling (LM), a technique commonly used in natural language processing, to learn the linear combinations of XML nodes that could occur within a lexical entry, and then evaluates each of these lexical entries against the learned patterns, looking for entries that diverge from the norm.

Technical users of XML often utilize tools to check the well-formedness of an XML document, or to determine the validity of a document as applied to a particular data schema. These help catch certain types of errors, such as syntax, or data relationship errors.

With many dictionary schemas, however, the structure within entries can vary from entry to entry. This structural permissiveness can allow a dictionary writer to introduce or underspecify ambiguous relationships, or to accidently place a node underneath an incorrect parent node in the entry's XML tree. These kinds of errors may be valid XML and may conform to the data specification, so they will not be caught by traditional XML tools, but they are *semantically* incorrect.

The LM technique described here linearizes the lexicon structure, ignoring the underlying text, converting the opening tags in XML into tokens, and then considering the string of tokens representing a lexical entry to be a sentence. A probabilistic language model is learned from these example sentences, and then that model is evaluated against each lexical entry in the corpus. Nodes that are in unusual positions produce a high perplexity, identifying possible anomaly points.

## 2. XML

Extensible Markup Language (XML), a text format used to store hierarchical data electronically, is often described by a data modeling definition such as a Document Type Definition (DTD), XML Schema (Gao, 2011), or RELAX NG (ISO, 2008). These data modeling definition documents use a regular language to define the data permissible in the XML document. Tools are available to validate, or ensure strict compliance of an XML document, to the data modeling definition. These tools result in a Boolean decision as to whether the data conforms to the specification, and are unable to alert the user to structurally valid, but illogical or rare structures that one may wish to investigate.

## 3. Structural Errors in Dictionaries

### 3.1 Dictionary Creation

Dictionaries are often the product of long-term research projects, or large-scale projects created quickly with multiple collaborators. Without strict conformance to a recording standard, entries can drift in style across time or between collaborators. Additionally, dictionaries can be large and complex, leaving them expensive to edit. Whether due to cost or deadlines, dictionaries are published that have errors and inconsistencies.

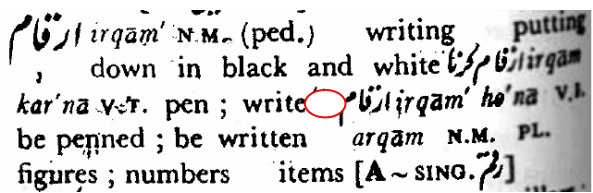

Figure 1: Missing Orthography. Scan from Qureshi (2003).

### 3.2 Dictionary Digitization

The process of digitizing dictionaries from a printed book by optical character recognition (OCR), or manually keying in content, can cause additional structural errors to be introduced. Typically in print dictionaries, typefaces, text size, text position, and unreserved symbols are used in combination to indicate the structure of a lexical entry and the scope of linguistic operators (such as English words *and* or *or*). Typographical errors that occur in the original print dictionary, misinterpretation by the OCR system, operator ambiguity, or typist error during the digitization stage can alter the intended structure of the dictionary. These errors can result in incorrect marking of subcomponents within a lexical entry or incorrectly understanding scope within the language examples. In bilingual dictionaries, translations may be forgotten (Figure 1), and languages may be mixed with no delineation (Figure 2).

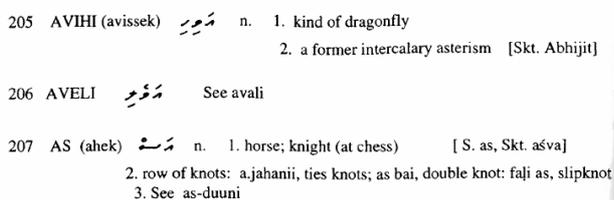

Figure 2: No signal between Dhivehi pronunciation and English meanings. Scan from Reynolds (2003).

### 4. Anomaly Detection Using Language Modeling

While language modeling is not a common approach for structural anomaly detection, it has been employed to detect anomalies in language use. Language models in natural language processing are commonly used to model linear combinations of word tokens or part-of-speech types. Lexicon XML structure is similar to the latter, in that the node names and attributes within the XML are chosen from a small closed class.

Xia & Wong (2006) used language modeling to tag lexemes in Chinese-language Internet chat transcripts as either standard Chinese, or anomalous. The authors noted that chat speak has a dynamic lexicon, and training corpora for supervised systems in this domain can obsolesce.

The authors trained trigram language models on standard Chinese newspaper corpora in order to induce typical values for trigram entropy on words and parts of speech in the standard language. They then learned a language model on a hybrid corpus consisting of newspaper corpora, and a chat transcript corpus. With the typical entropy values known from the newspaper corpus, they evaluated a language sample with the hybrid model. When a trigram had higher entropy than the standard average, they marked that trigram as an anomaly. The authors found words to be a better indicator of anomaly than POS tags, reaching an F-score of 0.85 for words and 0.70 for POS tags in their best conditions.

The authors do not qualify the data that is flagged anomalous. It would be interesting to know if this data is dialectal Chinese, misspelled words, bad grammar, emoticons, or lexemes unique to Chinese Internet chat. The POS tag condition is more comparable to our scenario, as both their POS categories and our structure description language have a small vocabulary.

Jabbari (2010) was interested in detection of anomalous words in the context of the words around them, which has practical applications including real-word spelling error detection and word sense disambiguation. The author examined an approach using bags of words, one using language models, and then a combination of the two. The language modeling approach marked a word as an anomaly if the probability of the word in that context was less than the expected probability of not having that word in the context. The language modeling system received an overall F-score of 0.71. It performed less well than the author's bag of words model, and the combined model. The author did not look at parts of speech, which is more relevant to our task.

### 5. LM Anomaly Detection for Flattened Structures

In the previous section, we showed how language modeling has been used to detect anomalies in a linear string of tokens. This section explains how to convert the XML tree into a linear string of tokens, and how this is used to build a language model.

The input to the language model is determined by specifying a repeating node in the XML file that contains child trees to be examined. Each of these repeating tree structures is traversed depth-first, and the element names and attributes of children are recorded to a buffer as a tag that identifies that element. In XML, a depth-first search is a linear scan of each node within a tree. At the end of each repeating node the buffer contains a single layer of whitespace-separated tags corresponding to a flattened representation of the tree. We call this a tag sentence. The tag sentences for all the repeating nodes form a corpus of tag sentences.

This corpus can then be used to train a statistical language model. For our experiments, we used the SRI Language Modeling Toolkit (SRILM) (Stolcke, 2002). SRILM includes command line programs and C++ libraries to calculate n-gram statistics for a language, and to measure the perplexity of a text sample to those statistics. SRILM reads and writes to a standard ARPA (Advanced Research Projects Agency) file format for n-gram models. There are other language modeling toolkits available. Typically, the differences between applications are in the speed of the evaluation, the size of the model created, or the statistical smoothing algorithms included for estimating low-occurrence combinations. In our case, speed and model size is not much of an issue, but estimating combinations of low token occurrence is. Since we are training and testing on the same dataset, advanced smoothing algorithms would not add any benefit. SRILM runs with Good-Turing and Katz back-off by default.

## 6. Evaluation

### 6.1 Dictionary and Evaluation Data

We perform our evaluations on a bilingual Urdu-English dictionary of 44,237 lexical entries (Qureshi, 2003). This dictionary has been edited by a team of linguists and computer scientists to remove errors using a change-tracking system we refer to as Dictionary Manipulation Language (DML) (Zajic et al., 2011). DML provides a number of benefits for dictionary editing, but the core advantage for this application is that DML can be used to mark every error discovered in the original source dictionary. From the change log, we can create a list of trees we know to be errorful that can be used for evaluating our automatic systems.

### 6.2 Tree Tiers

The entries in Qureshi (2003) can contain multiple senses, each of which can contain multiple word forms. An entry can also contain word forms directly. These are high-level structures within each entry that can vary significantly. In order to isolate where the errors occur within the entry, we partition some of the structure, performing evaluations on ENTRY, SENSE, and FORM nodes separately. For the ENTRY evaluation, the highest-level SENSE and FORM branches were collapsed into single nodes, with their descendants pruned. For the SENSE trees, descendent SENSE and FORM branches were collapsed. No branches were collapsed in the trees for FORM evaluation. We call the ENTRY, FORM, and SENSE trees *tiers*.

Our three tiers are listed in Table 1, showing the number of occurrences in the dictionary, as well as the number of nodes of that tree tier that had a hand-made correction within the tree.

| Tier | Count in Dictionary | Hand-corrected |
|---|---|---|
| ENTRY | 15,808 | 7,511 |
| FORM | 51,105 | 9,845 |
| SENSE | 88,465 | 20,037 |

Table 1: Tree counts and manual correction counts

### 6.3 n-gram Models

For each of these tiers, content and closing tags were removed, and the trees flattened to form a tag sentence. These three corpora were used to train 2-, 3- and 4-gram language models, without smoothing.

| Tree | Unique Tokens | 2-gram | 3-gram | 4-gram |
|---|---|---|---|---|
| ENTRY | 21 | 178 | 395 | 667 |
| FORM | 7 | 25 | 44 | 51 |
| SENSE | 22 | 183 | 384 | 628 |

Table 2: Unique Token and n-gram grammar counts at each tree level.

This language model serves to provide prototype trees for comparison, storing which tags can co-occur with which others, and what the likelihood of that co-occurrence will be. Table 2 lists the three tiers, the count of unique tokens (XML descendants) under that tier, and the number of unique n-grams created by the linear combination of those tokens.

### 6.4 Applying the models

Each tag sentence from the dictionary is then evaluated with this language model, producing statistical measurements for each flattened tree structure: log probability of the sentence (LOGPROB), average perplexity per word (PPW), and average perplexity per word with end tags (PPWET). LOGPROB and PPWET both evaluate trees against n-grams that contain START OF SENTENCE and END OF SENTENCE tags. This helps model differences between tokens that appear initial or final in the tag sentence.

We rank these measurements to force the trees into a decreasing order of anomalousness. For LOGPROB, the trees are sorted in ascending order, and for both PPW and PPWET, the trees are sorted in descending order.

For evaluation, we provide precision at the top R anomalies, where R can be {15, 30, 50, 100, 500, or 1000}. A hit occurs where a tree in the top R of our list has shown up in our errorful tree list. Precision at Rank is defined as the number of hits divided by R.

### 6.5 4-gram Results

Out of the three n-gram lengths evaluated, 4-grams performed the worst overall. The average of the six precisions at rank scores for each tree tier and each language model measurement were lower than those for both 3- and 2-grams. Several trials in the group did reach the best scores for their Tier-R combination, but these are matched in the 2- and 3-gram models. Results can be seen in Tables 3, 4, and 5.

| Tier / R | 15 | 30 | 50 | 100 | 500 | 1000 | AVG |
|---|---|---|---|---|---|---|---|
| ENTRY | .93 | .80 | .70 | .63 | .61 | .62 | .72 |
| FORM | .93 | .93 | .96 | .98 | .98 | **.99** | .96 |
| SENSE | **.93** | .93 | .92 | .89 | .61 | .56 | .81 |

Table 3: Descending PPWET 4-grams

| Tier / R | 15 | 30 | 50 | 100 | 500 | 1000 | AVG |
|---|---|---|---|---|---|---|---|
| ENTRY | **1.0** | .70 | .66 | .60 | .65 | .67 | .71 |
| FORM | .93 | .97 | .98 | **.99** | **.99** | **.99** | .98 |
| SENSE | **.93** | .93 | .92 | .90 | .56 | .54 | .80 |

Table 4: Descending PPW 4-grams

| Tier / R | 15 | 30 | 50 | 100 | 500 | 1000 | AVG |
|---|---|---|---|---|---|---|---|
| ENTRY | .87 | **.93** | .94 | .90 | .80 | .76 | .87 |
| FORM | .80 | .90 | .92 | .95 | .98 | .78 | .89 |
| SENSE | **.93** | .93 | .96 | .91 | .85 | .81 | .90 |

Table 5: Ascending LOGPROB 4-grams

### 6.6 3-gram Results

The 3-gram language model performed well, capturing the best average Tier / R trials for FORM with the PPW measurement. The results can be found in Tables 6, 7, and 8.

| Tier / R | 15 | 30 | 50 | 100 | 500 | 1000 | AVG |
|---|---|---|---|---|---|---|---|
| ENTRY | .93 | .83 | .80 | .73 | .69 | .72 | .78 |
| FORM | .93 | .93 | .96 | .98 | .98 | **.99** | .96 |
| SENSE | **.93** | .90 | .92 | .91 | .68 | .69 | .84 |

Table 6: Descending PPWET 3-grams

| Tier / R | 15 | 30 | 50 | 100 | 500 | 1000 | AVG |
|---|---|---|---|---|---|---|---|
| ENTRY | .93 | .73 | .72 | .69 | .69 | .74 | .75 |
| FORM | .97 | .98 | .99 | **.99** | **.99** | **.99** | **.99** |
| SENSE | **.93** | .93 | .92 | .91 | .64 | .59 | .82 |

Table 7: Descending PPW 3-grams

| Tier / R | 15 | 30 | 50 | 100 | 500 | 1000 | AVG |
|---|---|---|---|---|---|---|---|
| ENTRY | .87 | **.93** | .94 | **.93** | .86 | .84 | .90 |
| FORM | .87 | .93 | .94 | .95 | .98 | .78 | .91 |
| SENSE | **.93** | **.97** | **.98** | .94 | .91 | .87 | .93 |

Table 8: Ascending LOGPROB 3-grams

### 6.7 2-gram Results

2-gram language models results can be found in Tables 9, 10, and 11. This length n-gram performed the best, with the best average Tier / R trial for ENTRY and for SENSE using the LOGPROB measurement. This measurement, shown in Table 11, has the largest number of Tier/R trials with the highest precision.

| Tier / R | 15 | 30 | 50 | 100 | 500 | 1000 | AVG |
|---|---|---|---|---|---|---|---|
| ENTRY | .73 | .73 | .74 | .73 | .79 | .76 | .75 |
| FORM | .93 | .97 | .98 | .98 | .98 | **.99** | .97 |
| SENSE | **.93** | .83 | .74 | .81 | .74 | .78 | .81 |

Table 9: Descending PPWET 2-grams

| Tier / R | 15 | 30 | 50 | 100 | 500 | 1000 | AVG |
|---|---|---|---|---|---|---|---|
| ENTRY | .73 | .73 | .66 | .72 | .75 | .79 | .73 |
| FORM | .93 | .97 | .98 | **.99** | **.99** | **.99** | .98 |
| SENSE | **.93** | .93 | .92 | .83 | .71 | .76 | .85 |

Table 10: Descending PPW 2-grams

| Tier / R | 15 | 30 | 50 | 100 | 500 | 1000 | AVG |
|---|---|---|---|---|---|---|---|
| ENTRY | .87 | **.93** | **.96** | **.93** | **.91** | **.90** | **.92** |
| FORM | **1.0** | **1.0** | **1.0** | .97 | .98 | .78 | .96 |
| SENSE | **.93** | **.97** | **.98** | **.96** | **.96** | **.91** | **.95** |

Table 11: Ascending LOGPROB 2-grams

### 6.8 Other-grams

Unigram, 5-gram, and 6-gram models were also evaluated according to their LOGPROB. 5- and 6-gram models performed at a lower accuracy for nearly all R and tree levels. Unigram evaluations were inconclusive. Accuracy was slightly higher for some R, but some were far lower.

## 7. Conclusions

We presented a statistical error detection technique for dictionary structure that uses language modeling to rank anomalous dictionary trees for human review. To create the language model, we split the dictionary into three tiers-ENTRY, FORM, and SENSE, and flatten each to form a tag sentence. We create 2-, 3-, and 4-gram language models based on this flattened structure, and evaluate against the original dictionary using Perplexity Per Word (PPW), Perplexity Per Word with End Tags (PPWET), and log probability (LOGPROB). These measurements were ranked, and we presented Precision-at-Rank for all trials.

We found the highest precision Tier/R trials to be spread across several n-gram length language models, and several language model measurements. In general, we find that the best overall configuration is a 2-gram language model, which ranks the trees by ascending log probability. Averaging our six precisions for this metric, the system reached 92% precision on ENTRY error detection, 95% on SENSE, and 96% on FORM. Evaluating the top 50 anomalies, we reached 96% precision on ENTRY, 98% on SENSE, and 100% on FORM.

## 8. Comments

Though a large amount of man-hours were dedicated to the eradication of errors in our copy of the dictionary, we can make no assumption that we have found all of the errors present, and some of the trees that have not been marked bad, may indeed be errorful. Evaluation of our system, given this scenario, provides some difficulty. We have a small number of known-bad trees from the original source dictionary. The large remainder of trees is of questionable character, but are probably good. We cannot make large-scale automatic judgments on the questionable trees, but we can make sure the known-bad trees are ranked highly in our system. Actual precision should be considered at least the numbers reported. Unfortunately, without known-good trees, it is difficult to provide reliable recall measurements.

## 9. Future Work

### 9.1 Iterative language model improvement

As each error in a dictionary is corrected, the language model created from that dictionary improves. An iterative approach, having a linguist examine a small R, correcting the trees, and then rerunning the model, may be the most efficient use of a linguist's time.

### 9.2 Bootstrapping a cleaner model

With DML, we can find a small percentage of trees that are guaranteed to have had human review at some level. These trees are more likely to be correct than the completely untouched trees, and a corpus of the trees from the final dictionary could be used to create a higher quality language model to compare against the source dictionary.

### 9.3 Node-level anomaly detection

Evaluation of a language model on a tag sentence outputs a probability at each word. It would be interesting to show whether the peaks of perplexity correspond to the precise errors corrected in our dictionary.

### 9.4 Related systems

The language modeling approach is the first in a series of experiments examining anomaly detection on dictionary structure. We have several other frameworks currently under development, and expect approaches that harness structure, instead of flattening structure, will perform with higher accuracy. Additionally, we are planning work on a graphical tool to enable dictionary editors to interact with these anomaly detection systems, and plan to research how these systems can incorporate automatic error correction with assistance of an editor.

## 10. Acknowledgements

We would like to thank Mona Madgavkar for answering questions about the dictionary we were using for evaluations, as well as the other linguists that have contributed to this dictionary's correction. Additional thanks to C. Anton Rytting, Mike Maxwell, and two anonymous reviewers for their comments on this work.

This material is based upon work supported, in whole or in part, with funding from the United States Government. Any opinions, findings and conclusions, or recommendations expressed in this material are those of the author(s) and do not necessarily reflect the views of the University of Maryland, College Park and/or any agency or entity of the United States Government. Nothing in this report is intended to be and shall not be treated or construed as an endorsement or recommendation by the University of Maryland, United States Government, or the authors of the product, process, or service that is the subject of this report. No one may use any information contained or based on this report in advertisements or promotional materials related to any company product, process, or service or in support of other commercial purposes.